\documentclass[sigconf]{acmart}
\AtBeginDocument{%
  \providecommand\BibTeX{{%
    \normalfont B\kern-0.5em{\scshape i\kern-0.25em b}\kern-0.8em\TeX}}}

\setcopyright{acmcopyright}
\copyrightyear{2018}
\acmYear{2018}
\acmDOI{10.1145/1122445.1122456}

\acmConference[Woodstock '18]{Woodstock '18: ACM Symposium on Neural
  Gaze Detection}{June 03--05, 2018}{Woodstock, NY}
\acmBooktitle{Woodstock '18: ACM Symposium on Neural Gaze Detection,
  June 03--05, 2018, Woodstock, NY}
\acmPrice{15.00}
\acmISBN{978-1-4503-XXXX-X/18/06}

\usepackage{algorithm}
\usepackage{algorithmicx}
\usepackage{algpseudocode}
\usepackage{amsmath}
\usepackage{multirow}
\usepackage{caption}
\usepackage{subfigure}
\usepackage{graphicx}

\begin{document}

\title{Self-Supervised Dynamic Graph Representation Learning via Temporal Subgraph Contrast}



\author{Linpu Jiang}
\affiliation{%
  \institution{Nanjing University of Posts and Telecommunications}
  \streetaddress{9 Wenyuan Road}
  \city{Nanjing}
  \country{China}}
\email{1020041111@njupt.edu.cn}

\author{Ke-Jia Chen}
\affiliation{%
	\institution{Nanjing University of Posts and Telecommunications}
	\streetaddress{9 Wenyuan Road}
	\city{Nanjing}
	\country{China}}
\email{chenkj@njupt.edu.cn}

\author{Jingqiang Chen}
\affiliation{%
	\institution{Nanjing University of Posts and Telecommunications}
	\streetaddress{9 Wenyuan Road}
	\city{Nanjing}
	\country{China}}
\email{cjq@njupt.edu.cn}

\renewcommand{\shortauthors}{Linpu Jiang, Ke-Jia Chen and Jingqiang Chen}

\begin{abstract}
	Self-supervised learning on graphs has recently drawn a lot of attention due to its independence from labels and its robustness in representation. Current studies on this topic mainly use static information such as graph structures but cannot well capture dynamic information such as timestamps of edges. Realistic graphs are often dynamic, which means the interaction between nodes occurs at a specific time. This paper proposes a self-supervised dynamic graph representation learning framework (DySubC), which defines a temporal subgraph contrastive learning task to simultaneously learn the structural and evolutional features of a dynamic graph. Specifically, a novel temporal subgraph sampling strategy is firstly proposed, which takes each node of the dynamic graph as the central node and uses both neighborhood structures and edge timestamps to sample the corresponding temporal subgraph. The subgraph representation function is then designed according to the influence of neighborhood nodes on the central node after encoding the nodes in each subgraph. Finally, the structural and temporal contrastive loss are defined to maximize the mutual information between node representation and temporal subgraph representation. Experiments on five real-world datasets demonstrate that (1) DySubC performs better than the related baselines including two graph contrastive learning models and four dynamic graph representation learning models in the downstream link prediction task, and (2) the use of temporal information can not only sample more effective subgraphs, but also learn better representation by temporal contrastive loss.
\end{abstract}



\maketitle

\section{Introduction}
Graph-structured data, such as social networks \cite{Chen18}, collaboration networks \cite{Newman404} and chemical molecular graphs \cite{RenjieLiao19}, are ubiquitous in the real world. They naturally represent entities and their relationships. Graph representation learning aims to transform nodes into low-dimensional dense embeddings that preserve attributive and structural features of the graph. The method of using graph neural networks (GNN) \cite{Petar18, FelixWu19, Kipf16, MengQu19} has recently drawn considerable attention and achieved excellent performance. However, the supervised or semi-supervised GNNs heavily rely on labels with high acquisition cost, which could lead to poor generalization and weak robustness under label-related adversarial attacks \cite{YixinLi21}. 

Self-supervised learning can effectively alleviate the above problems. By designing appropriate training tasks on the graph, the self-supervised model can learn a more generalized graph representation in the absence of labels. Most of the existing methods focus on learning representations from the structural perspective by contrasting graph elements of different scales, such as node and local subgraph contrasting \cite{jiao2020sub-graph}, and node and global graph contrasting \cite{William18,sankararaman20}. 

\begin{figure}[h]
	\centering
	\includegraphics[width=1.05\linewidth,height=0.6\linewidth]{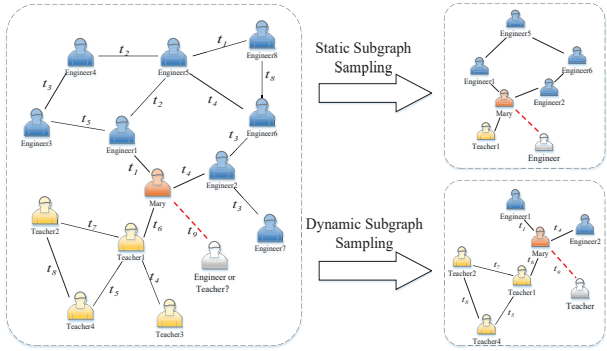}
	\caption{The influence of time-aware subgraphs on Mary's interaction at $t_9$ in social networks.}
	\label{introduction}
\end{figure}

However, neither graph sampling nor graph contrastive learning in these methods take the graph dynamics (especially the temporal information of edge generation) into account. As a result, the learned node representation cannot reflect the graph evolution. Take the social network in Figure \ref{introduction} as an example, where the edge represents the friend relationship of two person nodes. Each edge is marked with a timestamp ($t_1<t_2<…<t_9$), indicating the moment when they became friends. The central node Mary used to be an engineer, and most of her neighbor nodes were engineers before $t_6$. Recently, Mary changed her career to become a teacher and started to have teacher neighbor nodes (e.g., she will friend a teacher at time $t_9$). The static graph representation learning model samples subgraphs and encodes the nodes only based on structures, so it could predict that Mary will friend an engineer instead of a teacher at time $t_9$. By using temporal information, the dynamic subgraph sampling and node representation method can capture Mary’s relationship evolution, which may give a more accurate prediction.

Nodes are usually more closely related to their regional neighbors \cite{jiao2020sub-graph}. So intuitionally, frequent changes of neighbors will reflect potential changes of the central node and the latest neighbors that interacted with the central node are more valuable in node representation. This paper proposes a novel self-supervised dynamic graph representation learning method (DySubC), which learns node representation by sampling and contrasting with temporal subgraphs. To be specific, a temporal subgraph sampling strategy is firstly proposed, which takes each node as the central node to sample the corresponding temporal subgraph. The subgraphs are sampled by considering not only the local structure but also the temporal information of neighborhood interactions. Then, a certain GNN encoder is used to learn node representation for each temporal subgraph, thus output the representation of the central node and the summary of the temporal subgraph calculated by a time-aware readout function defined in this paper. Finally, the central node is paired with (a) its corresponding time-weighted subgraph as a positive sample, (b) any other temporal subgraph in a shuttled sampling set as a structural negative sample and (c) its corresponding unweighted subgraph as a temporal negative sample to train the whole model by maximizing the mutual information of the central node and its temporal subgraph.

In summary, the main contributions of the DySubC framework include:
\begin{itemize}
	\item {} This paper studies the problem of dynamic graph contrast learning for the first time and better captures the structural evolution characteristics of graphs by introducing temporal information.
	\item {} A temporal subgraph sampling method is proposed, which simultaneously uses the structural and temporal information of neighborhoods.
	\item {} A new readout function is defined to get the summary of a temporal subgraph, which takes the influence of the neighborhood nodes on the central node into account.
	\item {} A temporal subgraph contrast loss is defined including structure contrast loss and time contrast loss.
	\item {} Extensive experiments verify the superiority of DySubC in link prediction compared to the related continuous time graph representation learning models and graph contrastive learning models. The ablation study further proves the effectiveness of each time-enhanced module.
\end{itemize}

The rest of this paper includes the following sections. In Section \ref{RELATEDWORK}, the related work on dynamic graph representation learning and self-supervised graph learning is briefly reviewed. Section \ref{THE PROPOSED METHOD} introduces the proposed DySubC framework in detail. The experimental results are shown and analyzed in Section \ref{EXPERIMENTS}. Section \ref{CONCLUSION} concludes the paper.

\section{RELATED WORK}
\label{RELATEDWORK}

\subsection{Dynamic Graph Representation Learning}
Graph neural networks (GNNs) \cite{Petar18,FelixWu19, Kipf16,MengQu19} have achieved the competitive performance in static graph representation learning \cite{perozzi2014deepwalk,grover2016node2vec,tang2015line,wang2016structural}. Recently, the dynamic graph representation learning has drawn more and more attentions due to its ability to incorporate temporal information into representations. There are two different types of dynamic graphs, i.e., discrete time graphs and continuous time graphs. Accordingly, dynamic graph representation learning is also roughly divided into two categories.

Discrete time graph refers to a dynamic graph formalized as a series of multiple graph snapshots with the same time interval between them. This type of methods generally conduct representation learning on each snapshot first and then learns the evolution features of the graph structure over time through a sequence learning model (e.g., RNN \cite{zaremba2015recurrent}, self-attention \cite{vaswani2017attention}, etc.). Representative methods include DynGEM \cite{goyal2018dyngem}, DynamicTriad \cite{zhou2018dynamic}, EvolveGCN \cite{pareja2019evolvegcn}, DySAT \cite{sankar2020dysat} and so on. These methods focus more on capturing the global evolution features rather than the features of local continuous changes.

The continuous time graph is another formalization of dynamic graph where edges are marked with continuous timestamps. CTDNE \cite{nguyen2018continuous-time} proposes a temporal random walk method and then uses skip-gram to obtain node representation. HTNE \cite{zuo2018embedding} introduces the Hawkes process theory into the dynamic graph model and learns node representation based on the fact that the influence of neighbors on the central node will change over time. DyRep  \cite{trivedi2019dyrep} captures the interleaved dynamics of communication processes and correlation processes, thereby updating node representation. TGN \cite{rossi2020temporal} combines the memory module and graph-based operators to update node representations and improves the computational efficiency. The method proposed in this paper also falls into the category of continuous time graph representation learning, which mainly uses the changes of regional neighbors to learn node representations.

\subsection{Self-supervised Graph Learning}

Self-supervised learning uses pretext tasks to train the model with constructed supervised information from large-scale unsupervised data. It not only alleviates the problem of high cost of acquiring data labels, but also learns effective features. It has been successfully used in computer vision \cite{Jing20} and natural language processing \cite{Liu21}. For graph representation learning, DGI \cite{William18} is the first graph self-supervised learning method that uses the pretext task of maximizing mutual information between node representations and global graph representations. After that, a series of graph contrastive learning models emerged. Sankararaman et al. \cite{sankararaman20} propose a multi-view graph contrastive learning model MVGRL, which treats the original graph structure and graph diffusion as two different views to maximize mutual information between nodes and cross-view representations of large-scale graphs. The graph contrastive learning in GMI \cite{zhen2020graph} is achieved by maximizing the mutual information between the representation of each node and the original features of its one-hop neighbors. Sub-Con \cite{jiao2020sub-graph} maximizes the mutual information between node representations and subgraph representations, which can be used for large-scale graphs. 

In addition to the above methods, \cite{qiu2020gcc, yanqiao2020deep, zhu2021graph} introduce the contrastive learning method \cite{ting2020a} in machine vision into graph representation learning, which adopts different strategies for positive and negative sample construction and different loss functions. 

However, the existing graph self-supervised methods do not take into consideration the fact that graphs in real world are often dynamic. The temporal information is not well used not only in the pretext task but also in the definition of the objective function. The work of this paper attempts to bridge this gap.

\section{THE PROPOSED METHOD}
\label{THE PROPOSED METHOD}

\subsection{Preliminaries}
Before detailing the model, a formal definition of dynamic graph is given. Considering that the interaction between nodes occurs at a specific time and the graph is constantly changing over time, this paper models the dynamic graph as a continuous time graph.

\emph{\textbf{Definition 1} (Dynamic Graphs). Given a graph $G=(V,E_t,X)$, $V$ is a set of vertices, $E_t \subseteq V \times V \times \mathbb{R^+}$ is a set of edges with timestamp $t (t \subseteq \mathbb{R^+})$ and $X$ denotes the matrix of node attributes.  $e=(u,v,t) \subseteq E_t$ represents the interaction between node $u$ and $v$ at time $t$. Note that when the nodes are not attributed, one-hot encoding is often used to initialize $X$.}

\emph{\textbf{Problem 1} (Continuous-time dynamic graph representation learning). For a dynamic graph $G=(V,E_t,X)$, the task is to learn the mapping function $f: V \rightarrow  \mathbb{R}^D$ to embed the node in a $d$-dimensional vector space. The node representation is supposed to contain both structural and temporal information and suitable for downstream machine learning tasks such as link prediction.}

\subsection{Overview}
\begin{figure*}[h]
	\centering
	\includegraphics[width=\linewidth]{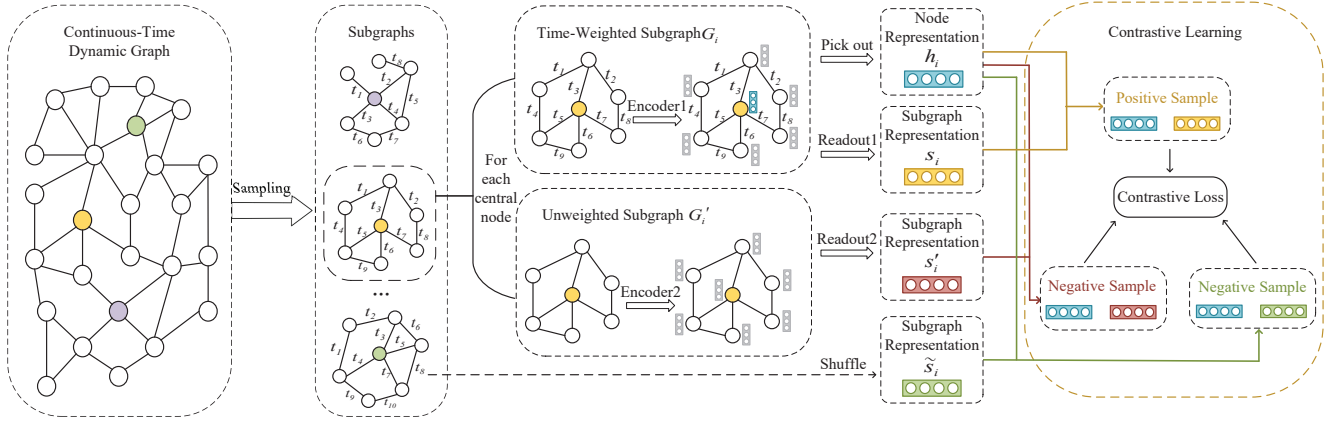}
	\caption{The overall framework of DySubC. DySubC first samples the temporal subgraph $G_i$ for each node. Then, taking the yellow central node as an example, DySubC encodes all the nodes in $G_i$ including the central node $i$ (represented as $h_i$). The time-weighted subgraph representation $s_i$ and the unweighted subgraph representation $s_i'$ are calculated by two readout functions respectively. Meanwhile, the other temporal subgraphs are shuffled to get a subgraph $\widetilde{G}_i$ with its representation $\widetilde{s}_i$. Finally, one positive sample and two negative samples are generated to calculate contrastive loss and train the model.}
	\label{framework}
\end{figure*}

The proposed DySubC (\textbf{Dy}namic graph representation learning via temporal \textbf{Sub}graph \textbf{C}ontrast) framework uses graph contrastive learning that captures the structural and temporal features from continuous-time graphs during the training phase without additional supervision. An overview of the DySubC framework is shown in Figure \ref{framework}, which mainly includes three time-enhanced modules. 

\begin{itemize}
	\item{\textbf{Temporal subgraph sampling.}} Firstly, a temporal subgraph for each node $i$ in the original graph is sampled using both structural and temporal information, generating a time-weighted subgraph $G_i=(X_i,A_i)$ and an unweighted counterpart $G_i'=(X_i',A_i')$. 
	\item{\textbf{Node and subgraph representation.}} Secondly, $G_i$ and $G_i'$ are encoded through GNN and then the summary of each subgraph is represented by a readout function, respectively. For each node $i$, we have its representation $h_i$, a time-weighted subgraph representation $s_i$ and an unweighted subgraph representation $s_i'$.  
	\item{\textbf{Temporal contrastive learning.}} Finally, a positive sample $s_i$, a temporal negative sample $s_i'$ and a structural negative sample $\widetilde{s}_i$ are constructed for each central node $i$ represented by $h_i$. The model is trained by maximizing the mutual information of the central node representation and the time-weighted subgraph representation. 
	
\end{itemize}

Note that the temporal subgraph sampling can be completed independently before the start of training. Thus, it does not take up the running time of the model. 

\subsection{Temporal Subgraph Sampling}

The temporal subgraph sampling module is first proposed for each node to generate training samples for self-supervised learning.

By considering both the structure of neighbors and the timestamp of edge interactions, the temporal subgraph sampler (see Algorithm \ref{alg::subsamping}) can sample a temporal subgraph with a fixed number of nodes $k$ for each central node $i$. The specific steps are as follows.

\renewcommand{\algorithmicrequire}{\textbf{Input:}}  
\renewcommand{\algorithmicensure}{\textbf{Output:}} 
\begin{algorithm}[h]
	\caption{Temporal Subgraph Sampler} 
	\label{alg::subsamping}
	\begin{algorithmic}[1]
		\Require
		Dynamic graph $G=(V,E_t,X)$;
		Subgraph size $k$.
		\Ensure
		A time-weighted subgraph $G_i=(X_i,A_i)$ and an unweighted subgraph $G_i'=(X_i',A_i')$ for each node $i$.
		
		\State Preprocess $G$ to get the time-weighted adjacency matrix $A$ and the unweighted adjacency matrix $A'$.
		\For{each node $i$}
		\State Initialize the queue $q=\emptyset$, the sampling pool $Pl=\emptyset$ and the number of sampled nodes $count$=0.
		\State Add $i$ to $q$ and $Pl$ respectively, count=1.
		
		\Repeat
		\State Add ${Neighbor(v) (\forall v \in q)}$ into the candidate set $Cand$.
		
		\If{ $|Cand| < (k-count)$}
		\State Add all the nodes in $Cand$ into $q$ and $Pl$, respectively.
		\State $count \gets count + |Cand|$.
		\Else 
		\State Calculate the importance score $S$ (Eq. \ref{1}-\ref{2}) of each node in the $Cand$.
		\State Take the $k-count$ nodes with the largest $S$ value from $Cand$ and add them into $Pl$. 
		\EndIf
		\Until{$count==k$}
		\State Calculate $G_i$ and $G_i'$ with $Pl$ using Eq. \ref{3}.
		\EndFor
		
	\end{algorithmic}
\end{algorithm}

Firstly, all first-order neighbors of $i$ are sampled. Since the number of first-order neighbors is usually less than $k$, the second-order or even the higher-order neighbors of $i$ may be sampled until the number of candidate nodes is greater than or equal to $k$. 

Then, if the number of candidate nodes exceeds $k$, a selection strategy is adopted to select more important candidate nodes into the sampling pool according to their \textit{importance score}, which is defined as the combination of \textit{structural importance score} $S_{structure}^j$ (Eq. \ref{1}) and \textit{temporal importance score} $S_{time}^j$ (described below), where $j$ is the node id in the sampling path.

The structural importance score $S_{structure}^j$ is simply defined as the degree of the node:
\begin{equation}
	S_{structure}^j=Degree(j).
	\label{1}
\end{equation}
Actually, $S_{structure}^j$ can also be defined using other measures such as the eigenvector centrality \cite{Mark2018} , influence in PageRank \cite{Mark2018}, etc.

The temporal importance score $S_{time}^j$ is a normalized largest timestamp of the edge connected to node $j$.

As a result, the importance score of node $j$ is denoted as:
\begin{equation}
	S^j=S_{time}^j+ \alpha S_{structure}^j,
	\label{2}
\end{equation}
where, $\alpha$ is the hyperparameter used to balance the influence of structure and time.

\begin{figure}[h]
	\centering
	\includegraphics[width=0.7\linewidth]{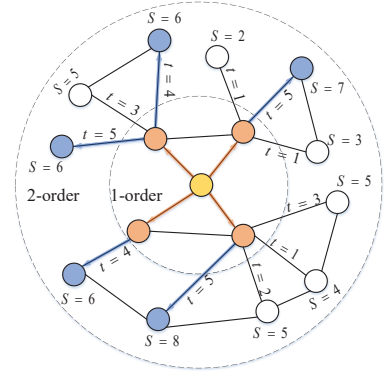}
	\caption{An illustrative example of temporal subgraph sampling.}
	\label{node_sampling}
\end{figure}

An example is shown in Figure \ref{node_sampling}, where $t$ represents the timestamp and $S$ represents the importance score. Assuming that $k=10$, for the yellow central node, its 4 first-order neighbors are first sampled and then its 11 second-order neighbors are also sampled, increasing to 16 nodes in the candidate set. According to Eq. \ref{2}, 5 blue nodes with the highest importance score will be saved as the sampling nodes.

Finally, after the $k$ nodes are sampled, the time-weighted subgraph $G_i$ and unweighted subgraph $G_i'$ of node $i$ are obtained referring to the original graph \cite{jiao2020sub-graph}, represented by the adjacency matrix $A_i$ and $A'_i$ respectively. The edge weight in $A_i$ is the normalization of its latest timestamp. The feature matrix $X_i$ and the adjacency matrix $A_i$ of $G_i$ are:

\begin{equation}
	X_i=X_{idx,:} , A_i=A_{idx,idx},
	\label{3}
\end{equation}
where, $idx$ represents the index of the sampled node. $X_{idx,:}$ is the row-wise (i.e. node-wise) indexed feature matrix. $A_{idx,idx}$ is the row-wise and col-wise indexed adjacency matrix. Similarly, the feature matrix $X_i'$ and adjacency matrix $A_i'$ of $G_i'$ are also obtained.

As a result, the output of the temporal subgraph sampling module for each node $i$ includes a time-weighted subgraph $G_i=(X_i,A_i)$ and an unweighted subgraph $G_i'=(X_i',A_i')$. Both of them will be used in the subsequent contrastive learning task.  

\subsection{Node and Subgraph Representation}
After obtaining $G_i=(X_i,A_i)$ and $G_i'=(X_i',A_i')$, the encoder $\mathcal{E}_1$ and $\mathcal{E}_2$ are used to output their representation $H_i$ and $H_i'$, respectively:
\begin{equation}
	H_i=\mathcal{E}_1(X_i,A_i),
	H_i'=\mathcal{E}_2(X_i',A_i').
	\label{4}
\end{equation}

For simplicity, both $\mathcal{E}_1$ and $\mathcal{E}_2$ adopt a one-layer graph convolutional network (GCN) \cite{Kipf16} that can efficiently aggregate neighbor information. Take $G_i$ as an example, the propagation rules are as follows:
\begin{equation}
	\mathcal{E}(X_i,A_i)=\sigma(\hat{D_i}^{-\frac{1}{2}}\hat{A_i}\hat{D_i}^{-\frac{1}{2}}X_iW),
	\label{5}
\end{equation}
where $\hat{A_i}=A_i+I$ is the adjacency matrix inserted into a self-loop, $\hat{D_i}$ is the corresponding degree matrix. The non-linear function $\sigma$ is the perametric ReLU (PReLU) function \cite{he2015delving} and $W$ is a learnable linear transformation. 

The representation $h_i$ of the central node $i$ is then picked out from the representation matrix $H_i$:
\begin{equation}
	h_i=\mathcal{P}(H_i),
	\label{6}
\end{equation}
where $\mathcal{P}$ denotes the pick-out operation.

To facilitate the subsequent contrast learning tasks, the time-weighted subgraph representation $s_i$ and the unweighted subgraph representation $s_i'$ are represented by using two readout functions $\mathcal{R}_1$ and $\mathcal{R}_2$, respectively.   

$\mathcal{R}_1$ is a time-aware readout function designed in this paper, which calculates the \textit{influence score} of any node $j$ on the central node $i$ (Eq. \ref{7}) and then performs a weighted average of node representations to get $s_i$ (Eq. \ref{8}). 

\begin{equation}
	Inf_j=\tau(i,j)+\beta\frac{1}{Dist(i,j)},
	\label{7}
\end{equation}
where, $\tau(i,j)$ represents the latest interaction timestamp between $i$ and $j$, $Dist(i, j)$ represents the shortest distance between $j$ and $i$ and $\beta$ is a hyperparameter. 

\begin{equation}
	s_i=\frac{1}{\sum_{j=1}^{k}Inf_j} \sum_{j=1}^{k}Inf_jh_j,
	\label{8}
\end{equation}
where, $k$ is the number of nodes in $G_i$.

$\mathcal{R}_2$ is used to simply average all node representations in the subgraph $G_i'$:
\begin{equation}
	s_i'=\frac{1}{k}\sum_{j=1}^{k}h_j',
	\label{9}
\end{equation}

\renewcommand{\algorithmicrequire}{\textbf{Input:}}  
\renewcommand{\algorithmicensure}{\textbf{Output:}} 
\begin{algorithm}[h]
	\caption{The DySubC algorithm} 
	\label{alg::model}
	\begin{algorithmic}[1]
		\Require
		Dynamic graph $G=(V,E_t,X)$.
		\Ensure
		node representations $\{h_1,h_2,...,h_{|V|}\}$
		\State Sample $\{G_1,G_2,...,G_{|V|}\}$ and $\{G_1',G_2',...,G_{|V|}'\}$ for each node by the temporal subgraph sampler.
		\While{not converge}
		\For{each $G_i$ and $G_i'$}
		\State Encode $G_i$ and $G_i'$ into $H_i$ and $H_i'$ through $\mathcal{E}_1$ and $\mathcal{E}_2$ (Eq. \ref{4}).
		\State Pick out $h_i=\mathcal{P}(H_i)$ (Eq. \ref{6}).
		\State Obtain $s_i$ and $s_i'$ through the readout function $\mathcal{R}_2$ and $\mathcal{R}_1$ respectively (Eq. \ref{8}-\ref{9}).
		\EndFor
		\State Shuffle the set of $\{s_1,s_2,...,s_{|V|}\}$ to generate $\{\widetilde{s}_1,\widetilde{s}_2,...,\widetilde{s}_{|V|}\}$.
		\State Construct a positive sample $s_i$, a temporal negative sample $s_i'$ and a structural negative sample $\widetilde{s}_i$. 
		\State Update parameters of $\mathcal{E}_1$ and $\mathcal{E}_2$ by Eq. \ref{13}.
		\EndWhile
		\For{each node $i$}
		\State Calculate the representation $h_i=\mathcal{P}(H_i)$ by Eq. \ref{6}.
		\EndFor
	\end{algorithmic}
\end{algorithm}

\subsection{Temporal Contrastive Learning}
Pretext tasks and the generation of positive and negative samples are crucial for self-supervised learning. As mentioned above, the dynamics of neighborhood nodes has greater influence on the central node than that of distant nodes. The pretext task is designed to make the central node strongly correlated to its regional neighbors. In our method, the mutual information of the central node and its corresponding temporal subgraph is maximized.

For the central node $i$, the pretext task is to contrast its real temporal subgraph to its fake temporal subgraph. Specifically, a positive sample, a structural negative sample and a temporal negative sample are constructed for $h_i$. The positive sample is the time-weighted subgraph representation $s_i$. The structural negative sample is generated by shuffling the representation set of time-weighted subgraphs, denoted as:

\begin{equation}
	\{\widetilde{s}_1,\widetilde{s}_2,...,\widetilde{s}_{|V|}\} = Shuffle(\{s_1,s_2,...,s_{|V|}\}).
	\label{10}
\end{equation}

The temporal negative sample is the unweighted subgraph representation $s_i'$, with the purpose to make the node representation $h_i$ closer to the time-weighted subgraph representation $s_i$ and farther away from the unweighted subgraph representation $s_i'$. The temporal contrast information is therefore emphasized.

Finally, the margin triplet loss is used to train the model as it is more favorable for subgraph contrastive learning \cite{jiao2020sub-graph}. The margin loss of the structural negative sample is defined as:
\begin{equation}
	\mathcal{L}_1=\frac{1}{|V|}\sum_{i=1}^{|V|}\mathbb{E}_{(X,A)}(-max(\sigma(h_is_i)-\sigma(h_i\widetilde{s}_i)+\phi,0)),
	\label{11}
\end{equation}
where $\sigma(x)=1/(1+exp(-x))$ is the sigmoid function and $\phi$ is the margin value. Similarly, the margin loss of the temporal negative sample is defined as:
\begin{equation}
	\mathcal{L}_2=\frac{1}{|V|}\sum_{i=1}^{|V|}\mathbb{E}_{(X,A)}(-max(\sigma(h_is_i)-\sigma(h_is_i')+\varphi,0)),
	\label{12}
\end{equation}
where $\varphi$ is the margin value. As a result, the total loss function of the model is:
\begin{equation}
	\mathcal{L}=\mathcal{L}_1+\lambda\mathcal{L}_2,
	\label{13}
\end{equation}
where $\lambda$ is a hyperparameter to balance two loss. 

The process of DySubC is summarized in Algorithm \ref{alg::model}.

\section{EXPERIMENTS}
\label{EXPERIMENTS}

The following experiments are conducted to evaluate our model from various aspects.
\begin{itemize}
	\item {} The performance of DynSubC and the related baseline methods in link prediction is compared, thereby reflecting their representation learning ability.
	\item {} The effectiveness of each proposed time-enhanced module and how it affects the overall model is evaluated.
	\item {} In the temporal subgraph sampling module, the impact of the subgraph size on the model is explored, and the balance between memory and performance is analyzed.
	\item {} The sensitivity of the model to hyperparameters is analyzed.
	\item {} The visualization of node representations obtained by DySubC and Sub-Con is compared.
\end{itemize}

Before detailing the experimental results and analysis, we start with a brief introduction of the datasets, the experimental settings and the baselines. 

\subsection{Datasets} For a comprehensive comparison, we use five widely used datasets collected from different types of real networks. The detailed statistics of the datasets are summarized in Table \ref{datasets}.

\begin{itemize}
	\item {} \textbf{fb-forum} \cite{opsahl2011triadic}. The dataset is a forum network similar to Facebook, obtained from online social networks. The directed edge $<u,v,t>$ means that user $u$ and user $v$ interacted at time $t$.
	\item {} \textbf{soc-sign-bitcoinalpha} \cite{kumar2016edge,kumar2018rev2}. This is a who-trusts-whom network among people who trade using Bitcoin on a platform called Bitcoin Alpha. The directed edge $<u,v,t>$ indicates that user $u$ trusted user $v$ at time $t$. 
	\item {} \textbf{ soc-wiki-elec} \cite{nr}. The dataset contains the administrator election and voting data based on the latest complete dump of Wikipedia page edit history (from January 3, 2008). The directed edge $<u,v,t>$ indicates that $u$ voted for $v$ at time $t$. 
	\item {} \textbf{ia-movielens-user2tags-10m} \cite{nr}. This bipartite network represents the tagging behaviors of MovieLens users. The nodes represent users and movies. The directed edge $<u,v,t>$ means that the user $u$ tagged the movie $v$ at $t$ time.
	\item {} \textbf{sx-mathoverflow-c2q} \cite{paranjape2017motifs}. It is a temporal interaction network on the stack exchange website Math Overflow. The directed edge $<u,v,t>$ means that user $u$ commented on user $v$'s question at time $t$.
	
\end{itemize}

\begin{table}
	\caption{Statistics of dynamic graph datasets.}
	\label{datasets}
	\resizebox{\columnwidth}{!}{
		\begin{tabular}{ccccc}
			\toprule
			Dynamic graph & $|V|$ & $|E_t|$ & Timespan (days) \\
			\midrule
			fb-forum & 899 & 33,720 & 164.49 \\
			soc-sign-bitcoinalpha & 3,783 & 24,186 & 1,901.00 \\
			soc-wiki-elec & 7,118 & 107,071 & 1,378.34 \\
			ia-movielens-user2tags-10m & 16,528 & 95,580 & 1,108.97 \\
			sx-mathoverflow-c2q & 16,836 & 203,639 & 2,349.00 \\
			
			\bottomrule
	\end{tabular}}
\end{table}

\begin{table*}
	\renewcommand\arraystretch{1.1}
	\caption{Performance of link prediction in terms of AUC score and accuracy. The best result is bolded.}
	\label{linkprediction}
	\begin{tabular}{ccccccccccc}
		\toprule[1pt]
		\multirow{2}{*}&
		\multicolumn{2}{c}{fb-forum} & \multicolumn{2}{c}{soc-sign-bitcoinalpha} &\multicolumn{2}{c}{soc-wiki-elec} 	&\multicolumn{2}{c}{ia-movielens-user2tags-10m}&\multicolumn{2}{c}{sx-mathoverflow-c2q} \cr
		\cmidrule(lr){2-11} & AUC &Accuracy& AUC &Accuracy& AUC &Accuracy& AUC &Accuracy& AUC &Accuracy \cr
		\midrule
		node2vec&	0.7351&	0.6703&	0.7001&	0.6514&	0.6877&	0.6439&	0.7118&	0.6477&	0.7521&0.6932 \\
		graphSAGE&	0.7465&	0.6815&	0.7389&	0.6887&	0.7454&	0.7014&	0.7648&	0.6994&	0.7863&0.7258 \\
		\cmidrule(lr){1-11}
		DGI&	0.8118&	0.7589&	0.7948&	0.7428&	0.8828&	0.8127&	0.8975&	0.8484&	0.8392&0.7803 \\
		Sub-Con&	0.8506&	0.7802&	0.8735&	0.8092&	0.8784&	0.8084&	0.8052&0.7321&		0.8281&0.7725 \\
		\cmidrule(lr){1-11}
		CTDNE&	0.7522&	0.6897&	0.7334& 0.6891&	0.7225&	0.6854&	0.7855&	0.7234&	0.8507 &0.7987 \\
		HTNE&	0.7756&	0.7023&	0.7365&	0.6872&	0.8253&0.7673&	0.8271&	0.7562&	0.8429&0.7904 \\
		DyRep&	0.7592&	0.6814&	0.8274&	0.7684&	0.8381&	0.7794&	0.8395&	0.7883&	0.8766&0.8237 \\
		TGN&	0.8678&0.7895&	0.8375&	0.7776&	0.8794&	0.8116&	0.8659&	0.8017&\textbf{	0.9371} &\textbf{0.8821}\\
		\cmidrule(lr){1-11}
		\textbf{DySubC}&	\textbf{0.8861} &\textbf{	0.8082}&	\textbf{0.9221}&	\textbf{0.8457}&	\textbf{0.9229}&\textbf{0.8558}&	\textbf{0.9524}&\textbf{0.8902}&	0.9302 &0.8791 \\
		\bottomrule[1pt]
	\end{tabular}
\end{table*}

\begin{table*}
	\renewcommand\arraystretch{1.1}
	\caption{Ablation studies. The best result performance is bolded.}
	\label{Ablation}

		\begin{tabular}{ccccccccccc}
			\toprule[1pt]
			\multirow{2}{*}&
			\multicolumn{2}{c}{fb-forum} & \multicolumn{2}{c}{soc-sign-bitcoinalpha} &\multicolumn{2}{c}{soc-wiki-elec} &\multicolumn{2}{c}{ia-movielens-user2tags-10m}&\multicolumn{2}{c}{sx-mathoverflow-c2q} \cr
			\cmidrule(lr){2-11} & AUC &Accuracy& AUC &Accuracy& AUC &Accuracy& AUC &Accuracy& AUC &Accuracy \cr
			
			\midrule
			DySubC$_{-S-N-R}$& 0.8442&0.7661&		0.8967&	0.8186&	0.8874&	0.8264&	0.9301&	0.8705&	0.8815& 0.8156 \\
			DySubC$_{-N-R}$& 0.8572&0.7782&		0.9149&	0.8388&	0.9124&	0.8462&	0.9395&	0.8751&	0.9017& 0.8458 \\
			DySubC$_{-R}$& 0.8821&0.8051&		0.9203&	0.8415&	0.9218&	0.8543&	0.9511&	0.8877&	0.9283&0.8768 \\
			\textbf{DySubC}& \textbf{0.8861}&\textbf{0.8082}&		\textbf{0.9221}&\textbf{0.8457}&		\textbf{0.9229}&\textbf{0.8558}&		\textbf{0.9524}&\textbf{0.8902}&		\textbf{0.9302} &\textbf{0.8791}\\
			\bottomrule[1pt]
	\end{tabular}
\end{table*}

\subsection{Experimental settings}

Since the nodes in the above datasets have no features, we use one-hot encoding as the initial features of the node. Considering that the downstream task is link prediction, we first sort the edges in the graph in ascending order of time, using the recent 25\% randomly divided as validation set (10\%) and test set (15\%) and the remaining 75\% as the training set. For the recent 25\% edges (i.e. positive samples), we randomly sample the same number of negative samples (unconnected node pairs).

For each experiment, the dimension of node representations is set to 128. A simple logistic regression classifier is trained and tested for link prediction using the embedding results. We train the model for 10 times on different data splits and report the average performance for fair evaluation. Consistent with previous work \cite{rossi2020temporal, trivedi2019dyrep, nguyen2018continuous-time, zuo2018embedding}, we also use AUC and accuracy indicator to evaluate the performance of link prediction.

In training, the Adam optimizer is used with an initial learning rate of 0.001. For all datasets, the size of subgraphs is set to 20. Both the margin value $\varphi$ and $\phi$ for the loss function are set to 0.75 and $\lambda$ is 0.5.

\subsection{Baselines} 
We use three types of representative baseline methods for comparison: (1) Continuous-time dynamic graph representation learning methods: TGN \cite{rossi2020temporal}, DyRep \cite{trivedi2019dyrep}, CTDNE \cite{nguyen2018continuous-time} and HTNE \cite{zuo2018embedding}; (2) State-of-art static graph self-supervised methods: DGI \cite{William18} and Sub-Con \cite{jiao2020sub-graph}; (3) Static graph representation learning benchmarks: Node2vec \cite{grover2016node2vec} and GraphSAGE \cite{hamilton2018inductive}. Note that when reproducing the codes of different models, we carefully select the reported optimal hyperparameters to ensure a fair comparison.

\subsection{Performance on Link Prediction}

The comparative results of all methods in link prediction performance are summarized in Table \ref{linkprediction}. Overall, our proposed model shows a competitive performance. Except that the AUC and accuracy of DySubC is slightly lower than that of TGN on the \textit{sx-mathoverflow-c2q} dataset, the model is superior to all baselines on the remaining datasets. The possible reason for the good performance of TGN on the \textit{sx-mathoverflow-c2q} dataset is that its memory module will be more advantageous for datasets with a large time span. The methods that use self-supervised learning (DGI, Sub-Con and DySubC) perform better than methods that do not use self-supervised learning (node2vec and graphSAGE).

More detailed observations and analysis are as follows. Firstly, the performance of our model is significantly higher than static self-supervised graph representation learning methods (i.e. DGI and Sub-Con). Especially on the \textit{sx-mathoverflow-c2q} dataset, the AUC score of DySubC is nearly 0.9 higher than DGI and 0.1 higher than Sub-Con. It verifies the importance of temporal information for graph representation learning models and the temporal subgraph contrastive method in this paper can effectively capture temporal information. Secondly, on the bipartite graph dataset (i.e., \textit{ia-movielens-user2tags-10m}), the performance of most continuous-time dynamic graph representation learning methods is poor. In contrast, the DySubC model achieves the best performance, which indicates that our method is more robust and can be applied to different types of graphs. Thirdly, DySubC performs significantly better than other models on \textit{soc-sign-bitcoinalpha} and \textit{ia-movielens-user2tags-10m} datasets. These two graphs are relatively sparse, which could cause the deterioration in model performance. However, DySubC may alleviate the above problem caused by the graph sparsity since it learns at the sub-graph level. Finally, the performance of Sub-Con on the \textit{ia-movielens-user2tags-10m} dataset is poor, probably because its subgraph sampling strategy is a personalized PageRank algorithm \cite{Mark2018}. It may not sample sufficient first-order neighbors, which are particularly important for the central node in the bipartite graph. 

\subsection{Ablation Studies}
In order to further observe the impact of three time-enhanced modules (i.e., temporal subgraph sampling, time-weighted subgraph representation and temporal contrastive learning) of DySubC, we conduct a series of ablation experiments by replacing each of the three modules with the counterpart that does not utilize the time information. The subscript -\textit{S} represents that the model replaces the temporal subgraph sampling with a subgraph sampling that only uses structural information. The subscript -\textit{N} means that the model does not use the negative sample $s_i'$ for contrastive learning. The subscript -\textit{R} stands for replacing our designed readout function with a simple average function in time-weighted subgraph representation. It is worth noting that DySubC$_{-S-N-R}$ and Sub-Con \cite{jiao2020sub-graph} are not equivalent, as their subgraph sampling strategies are different.

The ablation results are listed in Table \ref{Ablation}. It verifies the effectiveness of  each time-enhanced module, which consistently improves the model performance on all datasets. Especially when considering the negative sample $s_i'$, the performance of the model is significantly improved (see the comparative results of DySubC$_{-R}$ and DySubC$_{-N-R}$). The combination of three modules achieves the best performance. On the \textit{sx-mathoverflow-c2q dataset}, the final DySubC model gains an improvement of 0.05 AUC score compared to the base model with no time-enhanced module enabled.

\subsection{Subgraph Size Analysis}
\begin{figure}[h]
	\centering
	\includegraphics[width=0.9\linewidth]{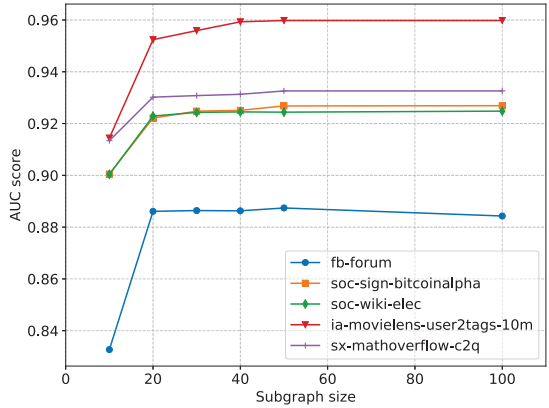}
	\caption{The impact of the subgraph size on DySubC.}
	\label{subgraph_size}
\end{figure}

This section studies the impact of the subgraph size in DynSubC on the five datasets. We adjust the subgraph size from 10 to 100 (including the central node), and the evaluation results are shown in Figure \ref{subgraph_size}. Note that in the experiment on \textit{ia-movielens-user2tags-10m} and \textit{sx-mathoverflow-c2q datasets}, the maximum subgraph size is set as 50 due to limited computational memory. The corresponding result with size 100 in the figure is approximated by the result with size 50, since the model performance tends to be stable.

As shown in the figure, the model achieves better performance when the size of the subgraph is larger. It is probably because neighborhood nodes contain more structural and temporal information, which helps to obtain a higher quality representation. However, in the \textit{fb-forum} dataset, the model with the subgraph size 100 performs worse than the model with the subgraph size 50. It may be because the dataset is small in size, a large size subgraph will contain nodes far away from the central node, which could be not beneficial to representation learning. When the size of subgraph is too small (e.g., 10), the performance of the model in all datasets is poor, which indicates that necessary information for learning is lost. When the subgraph size is 20, the model performs quite well on all five datasets and consumes less system memory than the model with a larger subgraph size.

\subsection{Sensitivity Analysis}
In this section, the sensitivity analysis is conducted on critical hyperparameters in DynSubC, i.e., $\alpha$ and $\beta$, which determine the quality of temporal subgraph sampling and time-weighted subgraph representation. Specifically, the value of $\alpha$ is increased from 2 to 20 with the step size 2 and the value of $\beta$ is increased from 0.4 to 3.6 with the step size 0.4. 

\begin{figure}[h]
	\centering
	\includegraphics[width=1.1\linewidth]{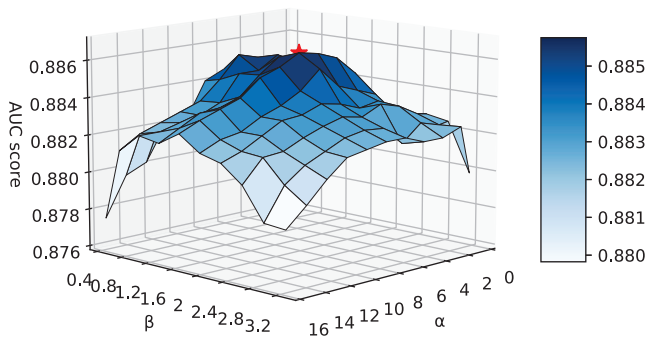}
	\caption{The performance of DySubC on the \textit{fb-forum} dataset with the change of two hyperparameters.}
	\label{parameter}
\end{figure}

The stability of the model under the perturbation of the two hyperparameters is observed. The results on the \textit{fb-forum} dataset are shown in Figure \ref{parameter}. The model performs best when $\alpha$ is 10 and $\beta$ is 1.6, both of which are around the median of their respective value ranges. It demonstrates that DynSubC is sensitive to these two hyperparameters. As we mentioned before, $\alpha$ is used to balance the structural and temporal information in the subgraph sampling. The result shows that the temporal information is at least as important as structural information when sampling and should be considered. $\beta$ is used to balance the influence of the latest interaction and its distance to the central node. Both are verified to be indispensable in subgraph representation. 

\subsection{Visualization Analysis}

\begin{figure}[htbp]
	\centering

	\subfigure[Sub-Con]{
			\centering
			\includegraphics[width=0.9\linewidth]{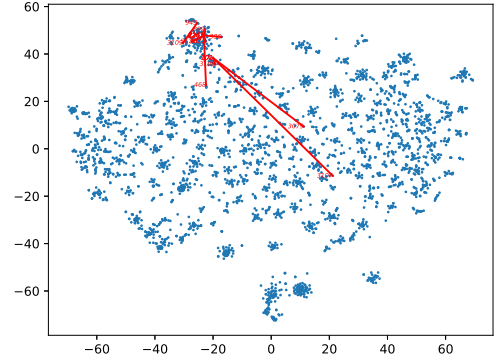}

	}%
	
	\subfigure[DySubC]{
		\centering
		\includegraphics[width=0.9\linewidth]{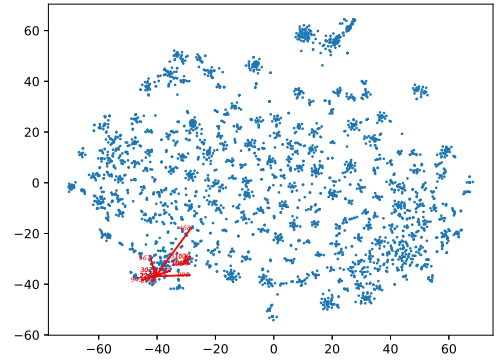}
	}%
	
	\centering
	\caption{Visualization comparison of node embedding and 10 latest interactions obtained by Sub-Con and DySubC on the \textit{soc-sign-bitcoinalpha} dataset.}
	\label{ksh}
\end{figure}

Finally, we use the tSNE algorithm \cite{maaten2008visualizing} to visualize the node representation and the latest interactions obtained by DySubC and Sub-Con. Figure \ref{ksh} shows the result on the \textit{soc-sign-bitcoinalpha} dataset. The red lines denote the latest 10 interactions. The visualization result shows that DySubC can better embed the evolutional features of nodes into the representation, so that the latest interacted nodes are also closer in the embedding space compared to Sub-Con, which is a static subgraph contrast method. Moreover, it is observed that the nodes that are recently active are more clustered in the visualization of DySubC than that of Sub-Con. It is consistent with the characteristics of the dataset, that is, users who trust each other are often in a community structure, so they are supposed to be closer in the embedding space.


\section{CONCLUSION}
\label{CONCLUSION}

In this paper, a novel self-supervised dynamic graph representation learning framework (DySubC) based on temporal subgraph contrast is proposed. The model learns the representation with both structural and temporal information by maximizing the mutual information of the node representation and its temporal subgraph representation. DySubC proposes three time-enhanced modules, which can not only sample more effective subgraphs, but also learn better representation by temporal contrast loss. The effectiveness of DySubC compared with related graph contrast learning methods and dynamic graph representation learning methods is demonstrated by empirical evaluation on multiple datasets. The success of DySubC provides an insight for the future study on continuous-time graph representation learning.

\bibliographystyle{ACM-Reference-Format}
\bibliography{sample-base}


\begin{thebibliography}{41}


\ifx \showCODEN    \undefined \def \showCODEN     #1{\unskip}     \fi
\ifx \showDOI      \undefined \def \showDOI       #1{#1}\fi
\ifx \showISBNx    \undefined \def \showISBNx     #1{\unskip}     \fi
\ifx \showISBNxiii \undefined \def \showISBNxiii  #1{\unskip}     \fi
\ifx \showISSN     \undefined \def \showISSN      #1{\unskip}     \fi
\ifx \showLCCN     \undefined \def \showLCCN      #1{\unskip}     \fi
\ifx \shownote     \undefined \def \shownote      #1{#1}          \fi
\ifx \showarticletitle \undefined \def \showarticletitle #1{#1}   \fi
\ifx \showURL      \undefined \def \showURL       {\relax}        \fi
\providecommand\bibfield[2]{#2}
\providecommand\bibinfo[2]{#2}
\providecommand\natexlab[1]{#1}
\providecommand\showeprint[2][]{arXiv:#2}

\bibitem[\protect\citeauthoryear{Chen, Ma, and Xiao}{Chen
  et~al\mbox{.}}{2018}]%
        {Chen18}
\bibfield{author}{\bibinfo{person}{Jie Chen}, \bibinfo{person}{Tengfei Ma},
  {and} \bibinfo{person}{Cao Xiao}.} \bibinfo{year}{2018}\natexlab{}.
\newblock \showarticletitle{FastGCN: Fast Learning with Graph Convolutional
  Networks via Importance Sampling}. In \bibinfo{booktitle}{\emph{6th
  International Conference on Learning Representations, {ICLR} 2018, Vancouver,
  BC, Canada, April 30 - May 3, 2018, Conference Track Proceedings}}.
  \bibinfo{publisher}{OpenReview.net}.
\newblock
\urldef\tempurl%
\url{https://openreview.net/forum?id=rytstxWAW}
\showURL{%
\tempurl}


\bibitem[\protect\citeauthoryear{Chen, Kornblith, Norouzi, and Hinton}{Chen
  et~al\mbox{.}}{2020}]%
        {ting2020a}
\bibfield{author}{\bibinfo{person}{Ting Chen}, \bibinfo{person}{Simon
  Kornblith}, \bibinfo{person}{Mohammad Norouzi}, {and}
  \bibinfo{person}{Geoffrey~E. Hinton}.} \bibinfo{year}{2020}\natexlab{}.
\newblock \showarticletitle{A Simple Framework for Contrastive Learning of
  Visual Representations}.
\newblock   \bibinfo{volume}{119} (\bibinfo{year}{2020}),
  \bibinfo{pages}{1597--1607}.
\newblock
\urldef\tempurl%
\url{http://proceedings.mlr.press/v119/chen20j.html}
\showURL{%
\tempurl}


\bibitem[\protect\citeauthoryear{Goyal, Kamra, He, and Liu}{Goyal
  et~al\mbox{.}}{2018}]%
        {goyal2018dyngem}
\bibfield{author}{\bibinfo{person}{Palash Goyal}, \bibinfo{person}{Nitin
  Kamra}, \bibinfo{person}{Xinran He}, {and} \bibinfo{person}{Yan Liu}.}
  \bibinfo{year}{2018}\natexlab{}.
\newblock \showarticletitle{DynGEM: Deep Embedding Method for Dynamic Graphs}.
\newblock \bibinfo{journal}{\emph{CoRR}}  \bibinfo{volume}{abs/1805.11273}
  (\bibinfo{year}{2018}).
\newblock
\showeprint[arXiv]{1805.11273}
\urldef\tempurl%
\url{http://arxiv.org/abs/1805.11273}
\showURL{%
\tempurl}


\bibitem[\protect\citeauthoryear{Grover and Leskovec}{Grover and
  Leskovec}{2016}]%
        {grover2016node2vec}
\bibfield{author}{\bibinfo{person}{Aditya Grover} {and} \bibinfo{person}{Jure
  Leskovec}.} \bibinfo{year}{2016}\natexlab{}.
\newblock \showarticletitle{node2vec: Scalable Feature Learning for Networks}.
\newblock  (\bibinfo{year}{2016}), \bibinfo{pages}{855--864}.
\newblock
\urldef\tempurl%
\url{https://doi.org/10.1145/2939672.2939754}
\showDOI{\tempurl}


\bibitem[\protect\citeauthoryear{Hamilton, Ying, and Leskovec}{Hamilton
  et~al\mbox{.}}{2017}]%
        {hamilton2018inductive}
\bibfield{author}{\bibinfo{person}{William~L. Hamilton},
  \bibinfo{person}{Zhitao Ying}, {and} \bibinfo{person}{Jure Leskovec}.}
  \bibinfo{year}{2017}\natexlab{}.
\newblock \showarticletitle{Inductive Representation Learning on Large Graphs}.
\newblock  (\bibinfo{year}{2017}), \bibinfo{pages}{1024--1034}.
\newblock
\urldef\tempurl%
\url{https://proceedings.neurips.cc/paper/2017/hash/5dd9db5e033da9c6fb5ba83c7a7ebea9-Abstract.html}
\showURL{%
\tempurl}


\bibitem[\protect\citeauthoryear{Hassani and Ahmadi}{Hassani and
  Ahmadi}{2020}]%
        {sankararaman20}
\bibfield{author}{\bibinfo{person}{Kaveh Hassani} {and} \bibinfo{person}{Amir
  Hosein~Khas Ahmadi}.} \bibinfo{year}{2020}\natexlab{}.
\newblock \showarticletitle{Contrastive Multi-View Representation Learning on
  Graphs}.
\newblock   \bibinfo{volume}{119} (\bibinfo{year}{2020}),
  \bibinfo{pages}{4116--4126}.
\newblock
\urldef\tempurl%
\url{http://proceedings.mlr.press/v119/hassani20a.html}
\showURL{%
\tempurl}


\bibitem[\protect\citeauthoryear{He, Zhang, Ren, and Sun}{He
  et~al\mbox{.}}{2015}]%
        {he2015delving}
\bibfield{author}{\bibinfo{person}{Kaiming He}, \bibinfo{person}{Xiangyu
  Zhang}, \bibinfo{person}{Shaoqing Ren}, {and} \bibinfo{person}{Jian Sun}.}
  \bibinfo{year}{2015}\natexlab{}.
\newblock \showarticletitle{Delving Deep into Rectifiers: Surpassing
  Human-Level Performance on ImageNet Classification}.
\newblock  (\bibinfo{year}{2015}), \bibinfo{pages}{1026--1034}.
\newblock
\urldef\tempurl%
\url{https://doi.org/10.1109/ICCV.2015.123}
\showDOI{\tempurl}


\bibitem[\protect\citeauthoryear{Jiao, Xiong, Zhang, Zhang, Zhang, and
  Zhu}{Jiao et~al\mbox{.}}{2020}]%
        {jiao2020sub-graph}
\bibfield{author}{\bibinfo{person}{Yizhu Jiao}, \bibinfo{person}{Yun Xiong},
  \bibinfo{person}{Jiawei Zhang}, \bibinfo{person}{Yao Zhang},
  \bibinfo{person}{Tianqi Zhang}, {and} \bibinfo{person}{Yangyong Zhu}.}
  \bibinfo{year}{2020}\natexlab{}.
\newblock \showarticletitle{Sub-graph Contrast for Scalable Self-Supervised
  Graph Representation Learning}.
\newblock  (\bibinfo{year}{2020}), \bibinfo{pages}{222--231}.
\newblock
\urldef\tempurl%
\url{https://doi.org/10.1109/ICDM50108.2020.00031}
\showDOI{\tempurl}


\bibitem[\protect\citeauthoryear{Jing and Tian}{Jing and Tian}{2021}]%
        {Jing20}
\bibfield{author}{\bibinfo{person}{Longlong Jing} {and} \bibinfo{person}{Yingli
  Tian}.} \bibinfo{year}{2021}\natexlab{}.
\newblock \showarticletitle{Self-Supervised Visual Feature Learning With Deep
  Neural Networks: {A} Survey}.
\newblock \bibinfo{journal}{\emph{{IEEE} Trans. Pattern Anal. Mach. Intell.}}
  \bibinfo{volume}{43}, \bibinfo{number}{11} (\bibinfo{year}{2021}),
  \bibinfo{pages}{4037--4058}.
\newblock
\urldef\tempurl%
\url{https://doi.org/10.1109/TPAMI.2020.2992393}
\showDOI{\tempurl}


\bibitem[\protect\citeauthoryear{Kipf and Welling}{Kipf and Welling}{2017}]%
        {Kipf16}
\bibfield{author}{\bibinfo{person}{Thomas~N. Kipf} {and} \bibinfo{person}{Max
  Welling}.} \bibinfo{year}{2017}\natexlab{}.
\newblock \showarticletitle{Semi-Supervised Classification with Graph
  Convolutional Networks}.
\newblock  (\bibinfo{year}{2017}).
\newblock
\urldef\tempurl%
\url{https://openreview.net/forum?id=SJU4ayYgl}
\showURL{%
\tempurl}


\bibitem[\protect\citeauthoryear{Kumar, Hooi, Makhija, Kumar, Faloutsos, and
  Subrahmanian}{Kumar et~al\mbox{.}}{2018}]%
        {kumar2018rev2}
\bibfield{author}{\bibinfo{person}{Srijan Kumar}, \bibinfo{person}{Bryan Hooi},
  \bibinfo{person}{Disha Makhija}, \bibinfo{person}{Mohit Kumar},
  \bibinfo{person}{Christos Faloutsos}, {and} \bibinfo{person}{V.~S.
  Subrahmanian}.} \bibinfo{year}{2018}\natexlab{}.
\newblock \showarticletitle{{REV2:} Fraudulent User Prediction in Rating
  Platforms}. In \bibinfo{booktitle}{\emph{Proceedings of the Eleventh {ACM}
  International Conference on Web Search and Data Mining, {WSDM} 2018, Marina
  Del Rey, CA, USA, February 5-9, 2018}},
  \bibfield{editor}{\bibinfo{person}{Yi~Chang}, \bibinfo{person}{Chengxiang
  Zhai}, \bibinfo{person}{Yan Liu}, {and} \bibinfo{person}{Yoelle Maarek}}
  (Eds.). \bibinfo{publisher}{{ACM}}, \bibinfo{pages}{333--341}.
\newblock
\urldef\tempurl%
\url{https://doi.org/10.1145/3159652.3159729}
\showDOI{\tempurl}


\bibitem[\protect\citeauthoryear{Kumar, Spezzano, Subrahmanian, and
  Faloutsos}{Kumar et~al\mbox{.}}{2016}]%
        {kumar2016edge}
\bibfield{author}{\bibinfo{person}{Srijan Kumar}, \bibinfo{person}{Francesca
  Spezzano}, \bibinfo{person}{V.~S. Subrahmanian}, {and}
  \bibinfo{person}{Christos Faloutsos}.} \bibinfo{year}{2016}\natexlab{}.
\newblock \showarticletitle{Edge Weight Prediction in Weighted Signed
  Networks}. In \bibinfo{booktitle}{\emph{{IEEE} 16th International Conference
  on Data Mining, {ICDM} 2016, December 12-15, 2016, Barcelona, Spain}},
  \bibfield{editor}{\bibinfo{person}{Francesco Bonchi}, \bibinfo{person}{Josep
  Domingo{-}Ferrer}, \bibinfo{person}{Ricardo Baeza{-}Yates},
  \bibinfo{person}{Zhi{-}Hua Zhou}, {and} \bibinfo{person}{Xindong Wu}} (Eds.).
  \bibinfo{publisher}{{IEEE} Computer Society}, \bibinfo{pages}{221--230}.
\newblock
\urldef\tempurl%
\url{https://doi.org/10.1109/ICDM.2016.0033}
\showDOI{\tempurl}


\bibitem[\protect\citeauthoryear{Liao, Zhao, Urtasun, and Zemel}{Liao
  et~al\mbox{.}}{2019}]%
        {RenjieLiao19}
\bibfield{author}{\bibinfo{person}{Renjie Liao}, \bibinfo{person}{Zhizhen
  Zhao}, \bibinfo{person}{Raquel Urtasun}, {and} \bibinfo{person}{Richard~S.
  Zemel}.} \bibinfo{year}{2019}\natexlab{}.
\newblock \showarticletitle{LanczosNet: Multi-Scale Deep Graph Convolutional
  Networks}.
\newblock  (\bibinfo{year}{2019}).
\newblock
\urldef\tempurl%
\url{https://openreview.net/forum?id=BkedznAqKQ}
\showURL{%
\tempurl}


\bibitem[\protect\citeauthoryear{Liu, Zhang, Hou, Wang, Mian, Zhang, and
  Tang}{Liu et~al\mbox{.}}{2020}]%
        {Liu21}
\bibfield{author}{\bibinfo{person}{Xiao Liu}, \bibinfo{person}{Fanjin Zhang},
  \bibinfo{person}{Zhenyu Hou}, \bibinfo{person}{Zhaoyu Wang},
  \bibinfo{person}{Li Mian}, \bibinfo{person}{Jing Zhang}, {and}
  \bibinfo{person}{Jie Tang}.} \bibinfo{year}{2020}\natexlab{}.
\newblock \showarticletitle{Self-supervised Learning: Generative or
  Contrastive}.
\newblock \bibinfo{journal}{\emph{CoRR}}  \bibinfo{volume}{abs/2006.08218}
  (\bibinfo{year}{2020}).
\newblock
\showeprint[arXiv]{2006.08218}
\urldef\tempurl%
\url{https://arxiv.org/abs/2006.08218}
\showURL{%
\tempurl}


\bibitem[\protect\citeauthoryear{Liu, Pan, Jin, Zhou, Xia, and Yu}{Liu
  et~al\mbox{.}}{2021}]%
        {YixinLi21}
\bibfield{author}{\bibinfo{person}{Yixin Liu}, \bibinfo{person}{Shirui Pan},
  \bibinfo{person}{Ming Jin}, \bibinfo{person}{Chuan Zhou},
  \bibinfo{person}{Feng Xia}, {and} \bibinfo{person}{Philip~S. Yu}.}
  \bibinfo{year}{2021}\natexlab{}.
\newblock \showarticletitle{Graph Self-Supervised Learning: {A} Survey}.
\newblock \bibinfo{journal}{\emph{CoRR}}  \bibinfo{volume}{abs/2103.00111}
  (\bibinfo{year}{2021}).
\newblock
\showeprint[arXiv]{2103.00111}
\urldef\tempurl%
\url{https://arxiv.org/abs/2103.00111}
\showURL{%
\tempurl}


\bibitem[\protect\citeauthoryear{Maaten and Hinton}{Maaten and Hinton}{2008}]%
        {maaten2008visualizing}
\bibfield{author}{\bibinfo{person}{van der~Laurens Maaten} {and}
  \bibinfo{person}{Geoffrey Hinton}.} \bibinfo{year}{2008}\natexlab{}.
\newblock \showarticletitle{Visualizing Data using t-SNE}.
\newblock \bibinfo{journal}{\emph{JOURNAL OF MACHINE LEARNING RESEARCH}}
  (\bibinfo{year}{2008}), \bibinfo{pages}{2579--2605}.
\newblock


\bibitem[\protect\citeauthoryear{Newman}{Newman}{2001}]%
        {Newman404}
\bibfield{author}{\bibinfo{person}{M.~E.~J. Newman}.}
  \bibinfo{year}{2001}\natexlab{}.
\newblock \showarticletitle{The structure of scientific collaboration
  networks}.
\newblock \bibinfo{journal}{\emph{Proceedings of the National Academy of
  Sciences}} \bibinfo{volume}{98}, \bibinfo{number}{2} (\bibinfo{year}{2001}),
  \bibinfo{pages}{404--409}.
\newblock
\showISSN{0027-8424}
\urldef\tempurl%
\url{https://doi.org/10.1073/pnas.98.2.404}
\showDOI{\tempurl}
\showeprint{https://www.pnas.org/content/98/2/404.full.pdf}


\bibitem[\protect\citeauthoryear{Newman}{Newman}{2018}]%
        {Mark2018}
\bibfield{author}{\bibinfo{person}{Mark E.~J. Newman}.}
  \bibinfo{year}{2018}\natexlab{}.
\newblock \bibinfo{booktitle}{\emph{Networks: An Introduction (Second
  Edition)}}.
\newblock \bibinfo{publisher}{Oxford University Press}.
\newblock


\bibitem[\protect\citeauthoryear{Nguyen, Lee, Rossi, Ahmed, Koh, and
  Kim}{Nguyen et~al\mbox{.}}{2018}]%
        {nguyen2018continuous-time}
\bibfield{author}{\bibinfo{person}{Giang~Hoang Nguyen},
  \bibinfo{person}{John~Boaz Lee}, \bibinfo{person}{Ryan~A. Rossi},
  \bibinfo{person}{Nesreen~K. Ahmed}, \bibinfo{person}{Eunyee Koh}, {and}
  \bibinfo{person}{Sungchul Kim}.} \bibinfo{year}{2018}\natexlab{}.
\newblock \showarticletitle{Continuous-Time Dynamic Network Embeddings}.
\newblock  (\bibinfo{year}{2018}), \bibinfo{pages}{969--976}.
\newblock
\urldef\tempurl%
\url{https://doi.org/10.1145/3184558.3191526}
\showDOI{\tempurl}


\bibitem[\protect\citeauthoryear{Opsahl}{Opsahl}{2013}]%
        {opsahl2011triadic}
\bibfield{author}{\bibinfo{person}{Tore Opsahl}.}
  \bibinfo{year}{2013}\natexlab{}.
\newblock \showarticletitle{Triadic closure in two-mode networks: Redefining
  the global and local clustering coefficients}.
\newblock \bibinfo{journal}{\emph{Soc. Networks}} \bibinfo{volume}{35},
  \bibinfo{number}{2} (\bibinfo{year}{2013}), \bibinfo{pages}{159--167}.
\newblock
\urldef\tempurl%
\url{https://doi.org/10.1016/j.socnet.2011.07.001}
\showDOI{\tempurl}


\bibitem[\protect\citeauthoryear{Paranjape, Benson, and Leskovec}{Paranjape
  et~al\mbox{.}}{2017}]%
        {paranjape2017motifs}
\bibfield{author}{\bibinfo{person}{Ashwin Paranjape},
  \bibinfo{person}{Austin~R. Benson}, {and} \bibinfo{person}{Jure Leskovec}.}
  \bibinfo{year}{2017}\natexlab{}.
\newblock \showarticletitle{Motifs in Temporal Networks}.
\newblock  (\bibinfo{year}{2017}), \bibinfo{pages}{601--610}.
\newblock
\urldef\tempurl%
\url{https://doi.org/10.1145/3018661.3018731}
\showDOI{\tempurl}


\bibitem[\protect\citeauthoryear{Pareja, Domeniconi, Chen, Ma, Suzumura,
  Kanezashi, Kaler, Schardl, and Leiserson}{Pareja et~al\mbox{.}}{2020}]%
        {pareja2019evolvegcn}
\bibfield{author}{\bibinfo{person}{Aldo Pareja}, \bibinfo{person}{Giacomo
  Domeniconi}, \bibinfo{person}{Jie Chen}, \bibinfo{person}{Tengfei Ma},
  \bibinfo{person}{Toyotaro Suzumura}, \bibinfo{person}{Hiroki Kanezashi},
  \bibinfo{person}{Tim Kaler}, \bibinfo{person}{Tao~B. Schardl}, {and}
  \bibinfo{person}{Charles~E. Leiserson}.} \bibinfo{year}{2020}\natexlab{}.
\newblock \showarticletitle{EvolveGCN: Evolving Graph Convolutional Networks
  for Dynamic Graphs}.
\newblock  (\bibinfo{year}{2020}), \bibinfo{pages}{5363--5370}.
\newblock
\urldef\tempurl%
\url{https://aaai.org/ojs/index.php/AAAI/article/view/5984}
\showURL{%
\tempurl}


\bibitem[\protect\citeauthoryear{Peng, Huang, Luo, Zheng, Rong, Xu, and
  Huang}{Peng et~al\mbox{.}}{2020}]%
        {zhen2020graph}
\bibfield{author}{\bibinfo{person}{Zhen Peng}, \bibinfo{person}{Wenbing Huang},
  \bibinfo{person}{Minnan Luo}, \bibinfo{person}{Qinghua Zheng},
  \bibinfo{person}{Yu Rong}, \bibinfo{person}{Tingyang Xu}, {and}
  \bibinfo{person}{Junzhou Huang}.} \bibinfo{year}{2020}\natexlab{}.
\newblock \showarticletitle{Graph Representation Learning via Graphical Mutual
  Information Maximization}.
\newblock  (\bibinfo{year}{2020}), \bibinfo{pages}{259--270}.
\newblock
\urldef\tempurl%
\url{https://doi.org/10.1145/3366423.3380112}
\showDOI{\tempurl}


\bibitem[\protect\citeauthoryear{Perozzi, Al{-}Rfou, and Skiena}{Perozzi
  et~al\mbox{.}}{2014}]%
        {perozzi2014deepwalk}
\bibfield{author}{\bibinfo{person}{Bryan Perozzi}, \bibinfo{person}{Rami
  Al{-}Rfou}, {and} \bibinfo{person}{Steven Skiena}.}
  \bibinfo{year}{2014}\natexlab{}.
\newblock \showarticletitle{DeepWalk: online learning of social
  representations}.
\newblock  (\bibinfo{year}{2014}), \bibinfo{pages}{701--710}.
\newblock
\urldef\tempurl%
\url{https://doi.org/10.1145/2623330.2623732}
\showDOI{\tempurl}


\bibitem[\protect\citeauthoryear{Qiu, Chen, Dong, Zhang, Yang, Ding, Wang, and
  Tang}{Qiu et~al\mbox{.}}{2020}]%
        {qiu2020gcc}
\bibfield{author}{\bibinfo{person}{Jiezhong Qiu}, \bibinfo{person}{Qibin Chen},
  \bibinfo{person}{Yuxiao Dong}, \bibinfo{person}{Jing Zhang},
  \bibinfo{person}{Hongxia Yang}, \bibinfo{person}{Ming Ding},
  \bibinfo{person}{Kuansan Wang}, {and} \bibinfo{person}{Jie Tang}.}
  \bibinfo{year}{2020}\natexlab{}.
\newblock \showarticletitle{{GCC:} Graph Contrastive Coding for Graph Neural
  Network Pre-Training}.
\newblock  (\bibinfo{year}{2020}), \bibinfo{pages}{1150--1160}.
\newblock
\urldef\tempurl%
\url{https://doi.org/10.1145/3394486.3403168}
\showDOI{\tempurl}


\bibitem[\protect\citeauthoryear{Qu, Bengio, and Tang}{Qu
  et~al\mbox{.}}{2019}]%
        {MengQu19}
\bibfield{author}{\bibinfo{person}{Meng Qu}, \bibinfo{person}{Yoshua Bengio},
  {and} \bibinfo{person}{Jian Tang}.} \bibinfo{year}{2019}\natexlab{}.
\newblock \showarticletitle{{GMNN:} Graph Markov Neural Networks}.
\newblock   \bibinfo{volume}{97} (\bibinfo{year}{2019}),
  \bibinfo{pages}{5241--5250}.
\newblock
\urldef\tempurl%
\url{http://proceedings.mlr.press/v97/qu19a.html}
\showURL{%
\tempurl}


\bibitem[\protect\citeauthoryear{Rossi, Chamberlain, Frasca, Eynard, Monti, and
  Bronstein}{Rossi et~al\mbox{.}}{2020}]%
        {rossi2020temporal}
\bibfield{author}{\bibinfo{person}{Emanuele Rossi}, \bibinfo{person}{Ben
  Chamberlain}, \bibinfo{person}{Fabrizio Frasca}, \bibinfo{person}{Davide
  Eynard}, \bibinfo{person}{Federico Monti}, {and} \bibinfo{person}{Michael~M.
  Bronstein}.} \bibinfo{year}{2020}\natexlab{}.
\newblock \showarticletitle{Temporal Graph Networks for Deep Learning on
  Dynamic Graphs}.
\newblock \bibinfo{journal}{\emph{CoRR}}  \bibinfo{volume}{abs/2006.10637}
  (\bibinfo{year}{2020}).
\newblock
\showeprint[arXiv]{2006.10637}
\urldef\tempurl%
\url{https://arxiv.org/abs/2006.10637}
\showURL{%
\tempurl}


\bibitem[\protect\citeauthoryear{Rossi and Ahmed}{Rossi and Ahmed}{2015}]%
        {nr}
\bibfield{author}{\bibinfo{person}{Ryan~A. Rossi} {and}
  \bibinfo{person}{Nesreen~K. Ahmed}.} \bibinfo{year}{2015}\natexlab{}.
\newblock \showarticletitle{The Network Data Repository with Interactive Graph
  Analytics and Visualization}. In \bibinfo{booktitle}{\emph{Proceedings of the
  Twenty-Ninth {AAAI} Conference on Artificial Intelligence, January 25-30,
  2015, Austin, Texas, {USA}}}, \bibfield{editor}{\bibinfo{person}{Blai Bonet}
  {and} \bibinfo{person}{Sven Koenig}} (Eds.). \bibinfo{publisher}{{AAAI}
  Press}, \bibinfo{pages}{4292--4293}.
\newblock
\urldef\tempurl%
\url{http://www.aaai.org/ocs/index.php/AAAI/AAAI15/paper/view/9553}
\showURL{%
\tempurl}


\bibitem[\protect\citeauthoryear{Sankar, Wu, Gou, Zhang, and Yang}{Sankar
  et~al\mbox{.}}{2020}]%
        {sankar2020dysat}
\bibfield{author}{\bibinfo{person}{Aravind Sankar}, \bibinfo{person}{Yanhong
  Wu}, \bibinfo{person}{Liang Gou}, \bibinfo{person}{Wei Zhang}, {and}
  \bibinfo{person}{Hao Yang}.} \bibinfo{year}{2020}\natexlab{}.
\newblock \showarticletitle{DySAT: Deep Neural Representation Learning on
  Dynamic Graphs via Self-Attention Networks}.
\newblock  (\bibinfo{year}{2020}), \bibinfo{pages}{519--527}.
\newblock
\urldef\tempurl%
\url{https://doi.org/10.1145/3336191.3371845}
\showDOI{\tempurl}


\bibitem[\protect\citeauthoryear{Tang, Qu, Wang, Zhang, Yan, and Mei}{Tang
  et~al\mbox{.}}{2015}]%
        {tang2015line}
\bibfield{author}{\bibinfo{person}{Jian Tang}, \bibinfo{person}{Meng Qu},
  \bibinfo{person}{Mingzhe Wang}, \bibinfo{person}{Ming Zhang},
  \bibinfo{person}{Jun Yan}, {and} \bibinfo{person}{Qiaozhu Mei}.}
  \bibinfo{year}{2015}\natexlab{}.
\newblock \showarticletitle{{LINE:} Large-scale Information Network Embedding}.
\newblock  (\bibinfo{year}{2015}), \bibinfo{pages}{1067--1077}.
\newblock
\urldef\tempurl%
\url{https://doi.org/10.1145/2736277.2741093}
\showDOI{\tempurl}


\bibitem[\protect\citeauthoryear{Trivedi, Farajtabar, Biswal, and Zha}{Trivedi
  et~al\mbox{.}}{2019}]%
        {trivedi2019dyrep}
\bibfield{author}{\bibinfo{person}{Rakshit Trivedi}, \bibinfo{person}{Mehrdad
  Farajtabar}, \bibinfo{person}{Prasenjeet Biswal}, {and}
  \bibinfo{person}{Hongyuan Zha}.} \bibinfo{year}{2019}\natexlab{}.
\newblock \showarticletitle{DyRep: Learning Representations over Dynamic
  Graphs}.
\newblock  (\bibinfo{year}{2019}).
\newblock
\urldef\tempurl%
\url{https://openreview.net/forum?id=HyePrhR5KX}
\showURL{%
\tempurl}


\bibitem[\protect\citeauthoryear{Vaswani, Shazeer, Parmar, Uszkoreit, Jones,
  Gomez, Kaiser, and Polosukhin}{Vaswani et~al\mbox{.}}{2017}]%
        {vaswani2017attention}
\bibfield{author}{\bibinfo{person}{Ashish Vaswani}, \bibinfo{person}{Noam
  Shazeer}, \bibinfo{person}{Niki Parmar}, \bibinfo{person}{Jakob Uszkoreit},
  \bibinfo{person}{Llion Jones}, \bibinfo{person}{Aidan~N. Gomez},
  \bibinfo{person}{Lukasz Kaiser}, {and} \bibinfo{person}{Illia Polosukhin}.}
  \bibinfo{year}{2017}\natexlab{}.
\newblock \showarticletitle{Attention is All you Need}.
\newblock  (\bibinfo{year}{2017}), \bibinfo{pages}{5998--6008}.
\newblock
\urldef\tempurl%
\url{https://proceedings.neurips.cc/paper/2017/hash/3f5ee243547dee91fbd053c1c4a845aa-Abstract.html}
\showURL{%
\tempurl}


\bibitem[\protect\citeauthoryear{Velickovic, Cucurull, Casanova, Romero,
  Li{\`{o}}, and Bengio}{Velickovic et~al\mbox{.}}{2018}]%
        {Petar18}
\bibfield{author}{\bibinfo{person}{Petar Velickovic}, \bibinfo{person}{Guillem
  Cucurull}, \bibinfo{person}{Arantxa Casanova}, \bibinfo{person}{Adriana
  Romero}, \bibinfo{person}{Pietro Li{\`{o}}}, {and} \bibinfo{person}{Yoshua
  Bengio}.} \bibinfo{year}{2018}\natexlab{}.
\newblock \bibinfo{title}{Graph Attention Networks}.
\newblock
\newblock
\urldef\tempurl%
\url{https://openreview.net/forum?id=rJXMpikCZ}
\showURL{%
\tempurl}


\bibitem[\protect\citeauthoryear{Velickovic, Fedus, Hamilton, Li{\`{o}},
  Bengio, and Hjelm}{Velickovic et~al\mbox{.}}{2019}]%
        {William18}
\bibfield{author}{\bibinfo{person}{Petar Velickovic}, \bibinfo{person}{William
  Fedus}, \bibinfo{person}{William~L. Hamilton}, \bibinfo{person}{Pietro
  Li{\`{o}}}, \bibinfo{person}{Yoshua Bengio}, {and} \bibinfo{person}{R.~Devon
  Hjelm}.} \bibinfo{year}{2019}\natexlab{}.
\newblock \bibinfo{title}{Deep Graph Infomax}.
\newblock
\newblock
\urldef\tempurl%
\url{https://openreview.net/forum?id=rklz9iAcKQ}
\showURL{%
\tempurl}


\bibitem[\protect\citeauthoryear{Wang, Cui, and Zhu}{Wang
  et~al\mbox{.}}{2016}]%
        {wang2016structural}
\bibfield{author}{\bibinfo{person}{Daixin Wang}, \bibinfo{person}{Peng Cui},
  {and} \bibinfo{person}{Wenwu Zhu}.} \bibinfo{year}{2016}\natexlab{}.
\newblock \showarticletitle{Structural Deep Network Embedding}.
\newblock  (\bibinfo{year}{2016}), \bibinfo{pages}{1225--1234}.
\newblock
\urldef\tempurl%
\url{https://doi.org/10.1145/2939672.2939753}
\showDOI{\tempurl}


\bibitem[\protect\citeauthoryear{Wu, Jr., Zhang, Fifty, Yu, and Weinberger}{Wu
  et~al\mbox{.}}{2019}]%
        {FelixWu19}
\bibfield{author}{\bibinfo{person}{Felix Wu}, \bibinfo{person}{Amauri H.~Souza
  Jr.}, \bibinfo{person}{Tianyi Zhang}, \bibinfo{person}{Christopher Fifty},
  \bibinfo{person}{Tao Yu}, {and} \bibinfo{person}{Kilian~Q. Weinberger}.}
  \bibinfo{year}{2019}\natexlab{}.
\newblock \showarticletitle{Simplifying Graph Convolutional Networks}.
\newblock   \bibinfo{volume}{97} (\bibinfo{year}{2019}),
  \bibinfo{pages}{6861--6871}.
\newblock
\urldef\tempurl%
\url{http://proceedings.mlr.press/v97/wu19e.html}
\showURL{%
\tempurl}


\bibitem[\protect\citeauthoryear{Zaremba, Sutskever, and Vinyals}{Zaremba
  et~al\mbox{.}}{2014}]%
        {zaremba2015recurrent}
\bibfield{author}{\bibinfo{person}{Wojciech Zaremba}, \bibinfo{person}{Ilya
  Sutskever}, {and} \bibinfo{person}{Oriol Vinyals}.}
  \bibinfo{year}{2014}\natexlab{}.
\newblock \showarticletitle{Recurrent Neural Network Regularization}.
\newblock \bibinfo{journal}{\emph{CoRR}}  \bibinfo{volume}{abs/1409.2329}
  (\bibinfo{year}{2014}).
\newblock
\showeprint[arXiv]{1409.2329}
\urldef\tempurl%
\url{http://arxiv.org/abs/1409.2329}
\showURL{%
\tempurl}


\bibitem[\protect\citeauthoryear{Zhou, Yang, Ren, Wu, and Zhuang}{Zhou
  et~al\mbox{.}}{2018}]%
        {zhou2018dynamic}
\bibfield{author}{\bibinfo{person}{Le{-}kui Zhou}, \bibinfo{person}{Yang Yang},
  \bibinfo{person}{Xiang Ren}, \bibinfo{person}{Fei Wu}, {and}
  \bibinfo{person}{Yueting Zhuang}.} \bibinfo{year}{2018}\natexlab{}.
\newblock \showarticletitle{Dynamic Network Embedding by Modeling Triadic
  Closure Process}.
\newblock  (\bibinfo{year}{2018}), \bibinfo{pages}{571--578}.
\newblock
\urldef\tempurl%
\url{https://www.aaai.org/ocs/index.php/AAAI/AAAI18/paper/view/16572}
\showURL{%
\tempurl}


\bibitem[\protect\citeauthoryear{Zhu, Xu, Yu, Liu, Wu, and Wang}{Zhu
  et~al\mbox{.}}{2020}]%
        {yanqiao2020deep}
\bibfield{author}{\bibinfo{person}{Yanqiao Zhu}, \bibinfo{person}{Yichen Xu},
  \bibinfo{person}{Feng Yu}, \bibinfo{person}{Qiang Liu}, \bibinfo{person}{Shu
  Wu}, {and} \bibinfo{person}{Liang Wang}.} \bibinfo{year}{2020}\natexlab{}.
\newblock \showarticletitle{Deep Graph Contrastive Representation Learning}.
\newblock \bibinfo{journal}{\emph{CoRR}}  \bibinfo{volume}{abs/2006.04131}
  (\bibinfo{year}{2020}).
\newblock
\showeprint[arXiv]{2006.04131}
\urldef\tempurl%
\url{https://arxiv.org/abs/2006.04131}
\showURL{%
\tempurl}


\bibitem[\protect\citeauthoryear{Zhu, Xu, Yu, Liu, Wu, and Wang}{Zhu
  et~al\mbox{.}}{2021}]%
        {zhu2021graph}
\bibfield{author}{\bibinfo{person}{Yanqiao Zhu}, \bibinfo{person}{Yichen Xu},
  \bibinfo{person}{Feng Yu}, \bibinfo{person}{Qiang Liu}, \bibinfo{person}{Shu
  Wu}, {and} \bibinfo{person}{Liang Wang}.} \bibinfo{year}{2021}\natexlab{}.
\newblock \showarticletitle{Graph Contrastive Learning with Adaptive
  Augmentation}.
\newblock  (\bibinfo{year}{2021}), \bibinfo{pages}{2069--2080}.
\newblock
\urldef\tempurl%
\url{https://doi.org/10.1145/3442381.3449802}
\showDOI{\tempurl}


\bibitem[\protect\citeauthoryear{Zuo, Liu, Lin, Guo, Hu, and Wu}{Zuo
  et~al\mbox{.}}{2018}]%
        {zuo2018embedding}
\bibfield{author}{\bibinfo{person}{Yuan Zuo}, \bibinfo{person}{Guannan Liu},
  \bibinfo{person}{Hao Lin}, \bibinfo{person}{Jia Guo},
  \bibinfo{person}{Xiaoqian Hu}, {and} \bibinfo{person}{Junjie Wu}.}
  \bibinfo{year}{2018}\natexlab{}.
\newblock \showarticletitle{Embedding Temporal Network via Neighborhood
  Formation}.
\newblock  (\bibinfo{year}{2018}), \bibinfo{pages}{2857--2866}.
\newblock
\urldef\tempurl%
\url{https://doi.org/10.1145/3219819.3220054}
\showDOI{\tempurl}


\end{thebibliography}

\end{document}